\documentclass[sigconf]{acmart}
\usepackage{booktabs} 

\usepackage[utf8]{inputenc} 
\usepackage[T1]{fontenc}    
\usepackage{hyperref}       
\usepackage{url}            
\usepackage{booktabs}       
\usepackage{amsfonts}       
\usepackage{nicefrac}       
\usepackage{microtype}      
\usepackage{natbib}
\usepackage{graphicx}
\usepackage{lscape}
\usepackage{color}

\acmConference[]{}{}{} 
\acmYear{2018}

\begin{document}

\title{Combining Textual Content and Structure to Improve Dialog Similarity}

\author{Ana Paula Appel}
\affiliation{%
    \institution{IBM Research }}
\email{apappel@br.ibm.com}

\author{Paulo Cavalin}
\affiliation{%
    \institution{IBM Research }}
\email{pcavalin@br.ibm.com}

\author{Marisa Vasconcelos}
\affiliation{%
    \institution{IBM Research}}
\email{marisaav@br.ibm.com}

\author{Claudio Pinhanez}
\affiliation{%
    \institution{IBM Research}}
\email{csantosp@br.ibm.com}

\begin{abstract}
Chatbots, taking advantage of the success of the messaging apps and recent advances in Artificial Intelligence, have become very popular, from helping business to improve customer services to chatting to users for the sake of conversation and engagement (celebrity or personal bots). However, developing and improving a chatbot requires understanding their data generated by its users. Dialog data has a different nature of a simple question and answering interaction, in which context and temporal properties (turn order) creates a different understanding of such data. In this paper, we propose a novelty metric to compute dialogs' similarity based not only on the text content but also on the information related to the dialog structure. Our experimental results performed over the Switchboard dataset show that using evidence from both textual content and the dialog structure leads to more accurate results than using each measure in isolation.

\end{abstract}

\sloppy



\maketitle

\section{Introduction}




Chatbots have become more and more popular over the past years, owing for instance to the success of text messaging apps, advances in natural language processing techniques, and advances in scalability. Solutions based on chatbots allow industry services to be  24-hours a day available to their customers, cutting expenses, and automating many of their business processes~\cite{forbes}. Chatbots can be task-oriented, designed to a particular task to get information from the user to help complete the task (e.g., booking airline flights) or can be systems designed for extended conversations, mimicking of human-human interaction (e.g., psychological counseling)~\cite{chatbot_book}. A by-product of that is the increasing volume of data, in special textual dialogs, that is being generated by systems making use of chatbots. There is demand for approaches to better understanding such data to either gather insights from the users or to improve the chatbots since data analytics for chatbots is still an area that needs more investigation.


In such sense, one kind of analysis that could help understanding chatbot data is to find similarity between dialogs. Therefore, grouping similar dialogs can give not only insights about the customers that interact with the chatbot but also can help on the creation corpus-based chatbots that are built mining conversations between human-human or human-machine conversations~\cite{serban-2015-survey}. Finding similar objects is a traditional task in several areas as content-based image retrieval~\cite{liu2007survey}, text similarity identification~\cite{hotho2005brief}, code plagiarism detection~\cite{maurer2006plagiarism}, song identification~\cite{typke2005survey} and so on. For dialogs, such similarity-based retrieval could be useful to recover dialogs that present some similarity either in content or structure, and the identification of similar sets of dialogs can be valuable to better understand the data and further improve the content or structure of the chatbots.



Most of the work in dialog aims at studying specific details of the dialog or the users with the goal of, to cite a few, identify genders~\cite{kose2007mining}, classify speech acts~\cite{carpenter2011role,ferschke2012behind,schabus2016data}, identify socio-cultural phenomena~\cite{strzalkowski2010modeling}, understand user interaction \cite{khan2002mining}, study linguistic coordination
\cite{danescu2011chameleons}, or predict emotions~\cite{litman2004predicting}, \cite{maeireizo2004co}. Another front can be observed in the efforts to build corpora for dialog systems~\cite{serban2015survey}, dialog generation \cite{walker2012annotated,serban2017multiresolution,serban2017hierarchical}, dialog generation evaluation~\cite{liu2016not}, dialog control \cite{williams2016end} and build specific corpus such as Ubuntu~\cite{lowe2015ubuntu} or modeling Twitter as a dialog conversation~\cite{ritter2010unsupervised}. However, none of the supra-cited works address dialog similarity. A lot of works in dialog is focus on spoken dialog, to fill gaps~\cite{mesnil2017}, control dialog~\cite{williams2017hybrid, bordes2016learning, wen2016network} and measure sucess~\cite{su2015learning}.


One way to compare two dialogs is to solely consider the textual content of the dialog and then use some standard text similarity technique. Text similarity is a well-known task when people mine large volume of text data, and several techniques have been developed \cite{metzler2007similarity,gomaa2013survey}. Nevertheless, data from chatbots or dialogs present also a structure that goes beyond traditional text data. Chatbots data present some specific characteristics such as the order and the time interval of the utterances, the user interactions, speech acts, turns and so on. For that reason, computing dialog similarity becomes a complex task than traditional text similarity.




In this paper, we propose a similarity metric based on both textual content and structural properties of dialogs. To evaluate our proposed metric, we perform experiments on the well-know Switchboard dataset \cite{SWITCHBOARD92}. To the best of our knowledge, we are the first to address the problem of dialog similarity and thus, in this paper, we will present: 

\begin{itemize}
\item A new metric for detecting similarity between dialogs;
\item New metrics that combine metrics from dialog structure and metrics from the text to better represent dialogs; 
\item Experiments showing that these new metrics works better for dialogs them only text;
\item A discussion on the challenges of this new area and the next steps to be taken.
\end{itemize}

The remainder of this paper is organized as follows. The next section presents our proposed similarity metric. Section \ref{sec:eval} presents our evaluation methodology and reports the results of our experimental evaluation. Finally, Section \ref{sec:concl} concludes the paper and discuss future work.  









\section{Dialog Similarity} 
\label{sec:dialog}


Two dialogs can be similar considering different aspects. For instance, they can be similar in terms of textual content, that is, they address the same topics and/or the same entities, and thus forth. On the other hand, dialogs can also be similar in terms of structural features, such as the length of the dialog or the presence/absence of some kind of identified pattern of interaction. 

Our method combines evidences from both textual content and structural features to compute similarity between two dialogs. Considering these evidences as complementary, we combine them based on computing the similarity of content and structure individually and combine their similarity results with the Borda Count method at ranking level, so that we avoid the burden of dealing with different scales of the features or distances if combined differently.
.  

\subsection{Textual-based Similarity}

For computing the features related to similarity based on textual content, we simply convert the dialog to a free-form text, where each dialog utterance consists of a new sentence. Then, we apply standard text mining techniques, such as stop word removal methods and the computation of Term Frequency Inverse Document Frequency (TF-IDF) measure \cite{Weiss2012}, being defined as:  

\begin{equation}
w_(j)= tf(j) * \log_2(N/df(j))
\end{equation}

where $j$ is the $j-$th word in the dictionary, $\mbox{tf}(j)$ is its frequency in the documents, $N$ is the number of documents in the dataset,  and $\mbox{df}(j)$ is the number of documents in which the word appears.

Once those features are computed, the cosine distance is used to compute the similarity between the TF-IDF vector of two dialogs, as:
\begin{equation}
   \mbox{cosine}(d1, d2) = \sum(w_{d1}(j)*w_{d2}(j))/(norm(d1)*norm(d2)),
\end{equation}
where $\mbox{norm}(D) = \sqrt{\sum w(j)^2}$.

\subsection{Dialog Structural Similarity}

\begin{table}[htbp]
\centering
\begin{minipage}{.45\textwidth}
    \centering
    \small
    \begin{tabular}{|l|}\hline
        T1 - A: I would like to go from New York to Boston \\ 
        T2 - B: Where do you want to go from? \\
        T3 - A: I want to go from New York to Boston \\
        T4 - B: What is your origin and destination of your trip? \\
        T5 - A: I want to travel from New York to Boston \\
        T6 - B: Please, give your origin and destination. \\
       \hline
    \end{tabular}
    \caption{Example of a cycle in a dialog between two participants \textbf{A} and \textbf{B}.}
    \label{tab:cycle}
\end{minipage}
\end{table}

For the similarity based on structural features of the dialog, we considered information about the turns, in which each turn is the contribution of one a dialog participant. We also compute metrics related to conversation cycles. A cycle is a situation when one participant of the conversation has to rephrase his/her idea until it is clear for the other participant or he/she gives up on that topic. For example, if you have the following conversation to book a plane ticket (Table \ref{tab:cycle}).

The conversation above has the repetition of the same intent which is related to book a ticket. The participant has to rephrase three times the question, therefore there is a cycle of size four in the dialog. Our approach estimates the size of cycle computing the similarity of two turns, particularly we use cosine distance over the words in each turn.



In this way, using the measures described above we compute the following metrics for each dialog:

\begin{itemize}
\item Total number of turns;
\item Average number of words per turn;
\item Total number of cycles;
\item Average number of turns in a cycle. 
\end{itemize}

\begin{table}[hbtp]
\centering
    \begin{tabular}{|l|}\hline
        T1 - A: You know right away what you want. \\ 
        T2 - B: I know right away what we, what we want. \\ 
        T3 - B: I keep hearing about it. \\ 
        T4 - B: I keep hearing the advertisements of it. \\ 
        T5 - A: You can't find them. \\ 
        T6 - A: You can't find them. \\ 
        T7 - B: and, i thought i would really miss that. \\ 
        T8 - A: I would, too,\\ \hline
    \end{tabular}
    \caption{Dialog extract between participants \textbf{A} and \textbf{B}.}
    \label{tab:dialog1}
\end{table}

\begin{table}[hbtp]
    \centering
    \begin{tabular}{|l|}\hline
    Total number of turns: 8 \\ 
    Average number of words per turn: 5.9\\ 
    Total number of cycles: 3\\ 
    Average number of turns in cycle: 2\\ \hline
    \end{tabular}
    \caption{Metrics calculated from dialog of Table \ref{tab:dialog1}.}
    \label{tab:Metrics}
\end{table}

Table \ref{tab:dialog1} shows a dialog piece extracted from the Switchboard dataset \cite{SWITCHBOARD92} while Table \ref{tab:Metrics} summarizes our metrics for that example. 


\subsection{Combining Textual and Structural Metrics}

The combination of both type of metrics (i.e., textual content and dialog structure) is done using the Borda Count method, by considering the ranking of similarity for a given dialog. In greater detail, for each of the metrics, we first compute the distance matrix $D$, which is an $N \times N$ matrix where $N$ is total of dialogs, where position $d_{i,j} \in D$ contains the distance between dialogs $i$ and $j$. This matrix will be computed for the textual content metrics ($D_T$) and for the structural metrics ($D_S$). Then, for each row $i$ in $D$, the ranking matrix $R$, which is also $N \times N$, is computed. Each cell $r_{i,j} \in R$ contains the relative ranking of distance $d_{i,j}$ compared against all distances $d_{i,k}$, where $k \ne j$. We present Tables \ref{tab:distance_matrix} and \ref{tab:ranking_matrix} to illustrate this process. Considering that in the first row $d_{1,1} < d_{1,3} < d_{1,2}$, we assign cells $r_{1,1}$, $r_{1,3}$ and $r_{1,2}$ with 1, 2 and 3, respectively. The same process is repeated for all the other rows.

\begin{table}[htbp]
\centering
    \begin{tabular}{|ccc|}
    \hline
    0.0 & 0.2 & 0.1 \\
    0.2 & 0.0 & 0.3 \\
    0.1 & 0.3 & 0.0 \\
    \hline
    \end{tabular}
    \caption{An example for a distance matrix $D$.}
    \label{tab:distance_matrix}
\end{table}

\begin{table}[htbp]
    \centering
    \begin{tabular}{|ccc|}
    \hline
    1 & 3 & 2 \\
    2 & 1 & 3 \\
    2 & 3 & 1 \\
    \hline
    \end{tabular}
    \caption{An example for a ranking matrix $R$, computed from the distance matrix in Table~\ref{tab:distance_matrix}}
    \label{tab:ranking_matrix}
\end{table}

That said, for combining the results of two distance matrices denoted $D_T$ and $D_S$, and their corresponding ranking matrices $R_T$ and $R_S$, we need to simply compute a third distance matrix denoted $D_B$, which corresponds to the sum of matrices $R_T$ and $R_S$, and them normally compute matrix $R_B$.



\section{Evaluation} 
\label{sec:eval}
In this section, we describe the dataset used for our experiments and present the results of our experimental evaluation and comparison with the two other similarity metrics (e.g., only dialog textual content or only dialog structural features).

\subsection{Dataset} 
\label{sub:data}
We use the Switchboard dataset~\cite{SWITCHBOARD92} for evaluating our similarity metric. This dataset is composed by 1,154 dialogs, ranging from a minimum of 38 and a maximum of 509 turns. This dataset is one of the most influential spoken corpora, being applied on several different tasks. In this work, we consider the transcripts converted from the spoken dialogs to sequences of text turns.

\subsection{Experiments} 
\label{sub:expe}
As described in the previous section, our approach to compute the similarity between two dialogs combines both textual content features and dialog structural features. Since there is no labeled dataset for similar dialogs, we consider an unsupervised evaluation of the metrics.


Our evaluation works as follows. Considering the textual content features and the corresponding ranking matrix $R_T$, the structure features and the corresponding ranking matrix $R_S$, and matrix $R_B$ which is the combination of the other two matrices, we compute the mean squared error (MSE) between each matrix with the purpose of understanding how these matrices can be compared. These results are presented in Table~\ref{tab:distance_matrix_each}. We can observe that the MSE between $R_T$ and $R_S$ is much higher compared with that between $R_T$ and $R_B$ or between $R_S$ and $R_B$. This suggests that our combination method is able to incorporate into $R_B$ the information in both $R_T$ and $R_S$.

\begin{table}[htbp]
\centering
    \begin{tabular}{|c|ccc|}
    \hline
    R   & T & S & B \\
    \hline
    T & 0.0 & 221591 & 69178 \\
    S & 221591 & 0.0 & 68772 \\
    B & 69178 & 68772 & 0.0 \\
    \hline
    \end{tabular}
    \caption{Mean squared error between $R_T$, $R_S$, and $R_B$.}
    \label{tab:distance_matrix_each}
 \end{table}
 
\begin{table}[htbp]
    \centering
    \begin{tabular}{|c|ccc|}
    \hline
    R & (1) & (2) & (3) \\
    \hline
    Ordered matrix  & 242819 & 218111 & 232473 \\
    \hline
    \end{tabular}
    \caption{The MSE between the computed ranking matrices and a randomly defined ranking.}
    \label{tab:distance_matrix_base}
\end{table}

\begin{figure}[htbp]
  \centering
    \includegraphics[width=0.45\textwidth]{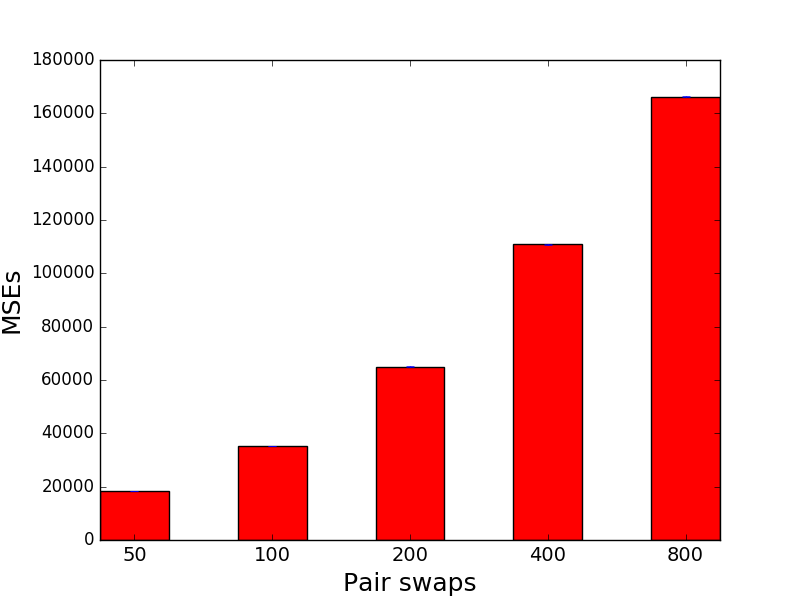}
    \caption{Mean Squared Error of different levels of perturbation of a randomly-defined matrix.}
    \label{fig:pair_swap_mses}
\end{figure}

For providing a better idea about the meaning of the computed MSE values, we have computed a random matrix then applied different degrees of perturbation in the ranking, and computed the MSE between the original and perturbed matrices. The perturbation consists of randomly picking a pair of elements and swapping their values. The degree of perturbation corresponds to the number of swaps that are performed. Figure~\ref{fig:pair_swap_mses} plots the MSEs of swapping from $50 \times 2^i$, with $i$ ranging from 1 to 5. With the values presented in Table~\ref{tab:distance_matrix_each}, we may say that the rankings in $R_T$ and $R_S$ are likely to be very distinct since the MSE between them is above that of swapping 800 pairs. On the other hand, the MSEs of $R_T$ and $R_B$, and that of $R_S$ and $R_B$, are between the swapping of 100 and 200 pairs (about 1/5 to 1/3 of the list). Thus, $R_B$ tends to present more similarity to $R_T$ and $R_S$, compared with the similarity of $R_T$ to $R_S$.

For a better understanding of the results, we have conducted a qualitative analysis of one dialog. We selected dialog \#90 (referred to as Original), and computed the most similar dialogs in terms of Content, which is dialog \#940, in terms of Structure, which is dialog \#354, and with the combination Content+Structure resulting in dialog \#85. Owning to space constraints, we do not present the contents of these dialog, but we present an analysis is terms of metrics for both content and structure.

In Table~\ref{tab:terms_intersection} we list the intersection of the most frequent terms in the dialogs considered as the most similar ones, in terms of Content, Structure, and Content+Structure, in our case study. We can observe that set of terms that intersect the original dialog and the most similar in terms of content, and the set that intersects the original dialog and the most similar in terms of structure, are quite different. Nevertheless, we observe that the terms considering the most similar dialog with Content+Structure cover most of the terms in the other two sets, showing that our combination method can be effective in combining both types of metrics. To complement this analysis, we also show that the intersection of terms among all dialogs is quite small, with only the terms 'people' and 'think', which also reinforces the potential of our approach.

\begin{table*}[htbp]
\centering
\begin{tabular}{|p{3cm}|p{8cm}|}
\hline
{\bf Dialogs Intersection} & {\bf Terms} \\
\hline
Original and Content & people, course, computer, long, research, problem, think \\
\hline
Original and Structure & people, recycling, paper, cans, bottles, plastic, program, location, collected, bins \\
\hline
Original and Content+Structure & people, recycling, paper, guy, bins, landfill, computer, plastic, collected, trash, separate, bottles, throw, school, cans \\
\hline
All & people, think \\
\hline
\end{tabular}
\caption{Intersection of terms between the most similar dialogs}
\label{tab:terms_intersection}
\end{table*}

In Table~\ref{tab:dialogMetrics} we present the values found for our structural features, for each of the dialogs. Not surprisingly, we observe that the Original dialog and the most similar one for Structure present very close values. On the other hand, the most similar dialog for Content present the most different values. And the most similar in Content+Structure does not present values as close as that of Structure to the Original dialog, but still the values are quite close if compared with the Content dialog.

\begin{table*}[htbp]
\centering
\begin{tabular}{|l|l|l|l|l|}\hline
\textbf{Dialogs} & $\#$ \textbf{turns} & \textbf{avg words per turn} & $\#$ \textbf{Cycle} & \textbf{avg turns per cycle} \\\hline
Original & 61 & 6.5 & 6 & 2.5 \\\hline
Structure & 65 & 6.5 & 5 & 2.6 \\\hline
Content &  108 & 4.2 & 9 & 6.9 \\\hline
Structure+Content & 75 & 5.7 & 6 & 2.8 \\\hline
\end{tabular}
    \caption{Structure Metrics calculated from dialogs used as example. The original and the most similar based on structure, content and structure and content respectively. }
    \label{tab:dialogMetrics}
\end{table*}

\section{Conclusion and Future Work} \label{sec:concl}
In this work, we tackle the problem of dialog similarity. For the best of our knowledge, we are the first to address the problem in the context that uses only the text is not enough to represent similarity between dialogs. 
We also presented a new metric for dialog similarity based on evidence from both textual content and the dialog structure using the Switchboard dataset.   
We found that considering structural properties of the dialog, such a number of turns and the present of cycles can improve the similarity detection accuracy when combine of textual content features (e.g., TF-IDF). 

As future work, we intend to analyze other temporal properties of the dialog such as precedence actions, other structural features such as speech acts, cycle size
and investigate the use of synonyms (e.g., wordvecs)  to capture meaning in the similarity. Finally, we would like to analyze the correlation between our similarity metric and human judgments. 


\small

\bibliographystyle{plain}
\bibliography{DialogSimilarity}

\begin{thebibliography}{10}

\bibitem{bordes2016learning}
Antoine Bordes and Jason Weston.
\newblock Learning end-to-end goal-oriented dialog.
\newblock {\em arXiv preprint arXiv:1605.07683}, 2016.

\bibitem{carpenter2011role}
Tamitha Carpenter and Emi Fujioka.
\newblock The role and identification of dialog acts in online chat.
\newblock In {\em Workshops at the Twenty-Fifth AAAI Conference on Artificial
  Intelligence}, 2011.

\bibitem{danescu2011chameleons}
Cristian Danescu-Niculescu-Mizil and Lillian Lee.
\newblock Chameleons in imagined conversations: A new approach to understanding
  coordination of linguistic style in dialogs.
\newblock In {\em Proceedings of the 2nd Workshop on Cognitive Modeling and
  Computational Linguistics}, pages 76--87. Association for Computational
  Linguistics, 2011.

\bibitem{ferschke2012behind}
Oliver Ferschke, Iryna Gurevych, and Yevgen Chebotar.
\newblock Behind the article: Recognizing dialog acts in wikipedia talk pages.
\newblock In {\em Proceedings of the 13th Conference of the European Chapter of
  the Association for Computational Linguistics}, pages 777--786. Association
  for Computational Linguistics, 2012.

\bibitem{SWITCHBOARD92}
J.~J. Godfrey, E.~C. Holliman, and J.~McDaniel.
\newblock Switchboard: telephone speech corpus for research and development.
\newblock In {\em [Proceedings] ICASSP-92: 1992 IEEE International Conference
  on Acoustics, Speech, and Signal Processing}, volume~1, pages 517--520 vol.1,
  Mar 1992.

\bibitem{gomaa2013survey}
Wael~H Gomaa and Aly~A Fahmy.
\newblock A survey of text similarity approaches.
\newblock {\em International Journal of Computer Applications}, 68(13), 2013.

\bibitem{hotho2005brief}
Andreas Hotho, Andreas N{\"u}rnberger, and Gerhard Paa{\ss}.
\newblock A brief survey of text mining.
\newblock In {\em Ldv Forum}, volume~20, pages 19--62, 2005.

\bibitem{chatbot_book}
Daniel Jurafsky and James~H. Martin.
\newblock Dialog systems and chatbots.
\newblock In {\em Speech and Language Processing: An Introduction to Natural
  Language Processing, Computational Linguistics, and Speech Recognition},
  chapter~29, pages 418--440. Prentice Hall PTR, 2000.

\bibitem{khan2002mining}
Faisal~M Khan, Todd~A Fisher, Lori Shuler, Tianhao Wu, and William~M Pottenger.
\newblock Mining chat-room conversations for social and semantic interactions.
\newblock {\em Computer Science and Engineering, Lehigh University}, 2002.

\bibitem{kose2007mining}
Cemal K{\"o}se, {\"O}zcan {\"O}zyurt, and Guychmyrat Amanmyradov.
\newblock Mining chat conversations for sex identification.
\newblock In {\em Pacific-Asia Conference on Knowledge Discovery and Data
  Mining}, pages 45--55. Springer, 2007.

\bibitem{litman2004predicting}
Diane~J Litman and Kate Forbes-Riley.
\newblock Predicting student emotions in computer-human tutoring dialogues.
\newblock In {\em Proceedings of the 42nd Annual Meeting on Association for
  Computational Linguistics}, page 351. Association for Computational
  Linguistics, 2004.

\bibitem{liu2016not}
Chia-Wei Liu, Ryan Lowe, Iulian~V Serban, Michael Noseworthy, Laurent Charlin,
  and Joelle Pineau.
\newblock How not to evaluate your dialogue system: An empirical study of
  unsupervised evaluation metrics for dialogue response generation.
\newblock {\em arXiv preprint arXiv:1603.08023}, 2016.

\bibitem{liu2007survey}
Ying Liu, Dengsheng Zhang, Guojun Lu, and Wei-Ying Ma.
\newblock A survey of content-based image retrieval with high-level semantics.
\newblock {\em Pattern recognition}, 40(1):262--282, 2007.

\bibitem{lowe2015ubuntu}
Ryan Lowe, Nissan Pow, Iulian Serban, and Joelle Pineau.
\newblock The ubuntu dialogue corpus: A large dataset for research in
  unstructured multi-turn dialogue systems.
\newblock {\em arXiv preprint arXiv:1506.08909}, 2015.

\bibitem{maeireizo2004co}
Beatriz Maeireizo, Diane Litman, and Rebecca Hwa.
\newblock Co-training for predicting emotions with spoken dialogue data.
\newblock In {\em Proceedings of the ACL 2004 on Interactive poster and
  demonstration sessions}, page~28. Association for Computational Linguistics,
  2004.

\bibitem{maurer2006plagiarism}
Hermann~A Maurer, Frank Kappe, and Bilal Zaka.
\newblock Plagiarism-a survey.
\newblock {\em J. UCS}, 12(8):1050--1084, 2006.

\bibitem{mesnil2017}
Grégoire Mesnil, Yann Dauphin, Kaisheng Yao, Yoshua Bengio, Li~Deng, Dilek
  Hakkani-Tür, Xiaodong He, Larry Heck, Gokhan Tur, Dong Yu, and Geoffrey
  Zweig.
\newblock Using recurrent neural networks for slot filling in spoken language
  understanding.
\newblock {\em IEEE/ACM Transactions on Audio, Speech, and Language
  Processing}, 23:530--539, 2015.

\bibitem{metzler2007similarity}
Donald Metzler, Susan Dumais, and Christopher Meek.
\newblock Similarity measures for short segments of text.
\newblock In {\em European Conference on Information Retrieval}, pages 16--27.
  Springer, 2007.

\bibitem{ritter2010unsupervised}
Alan Ritter, Colin Cherry, and Bill Dolan.
\newblock Unsupervised modeling of twitter conversations.
\newblock In {\em Human Language Technologies: The 2010 Annual Conference of
  the North American Chapter of the Association for Computational Linguistics},
  pages 172--180. Association for Computational Linguistics, 2010.

\bibitem{schabus2016data}
Dietmar Schabus, Brigitte Krenn, and Friedrich Neubarth.
\newblock Data-driven identification of dialogue acts in chat messages.
\newblock {\em Bochumer Linguistische Arbeitsberichte}, page 236, 2016.

\bibitem{serban-2015-survey}
Iulian~V. Serban, Ryan Lowe, Laurent Charlin, and Joelle Pineau.
\newblock A survey of available corpora for building data-driven dialogue
  systems.
\newblock {\em arXiv e-prints}, abs/1512.05742, December 2015.

\bibitem{serban2017multiresolution}
Iulian~Vlad Serban, Tim Klinger, Gerald Tesauro, Kartik Talamadupula, Bowen
  Zhou, Yoshua Bengio, and Aaron~C Courville.
\newblock Multiresolution recurrent neural networks: An application to dialogue
  response generation.
\newblock In {\em AAAI}, pages 3288--3294, 2017.

\bibitem{serban2015survey}
Iulian~Vlad Serban, Ryan Lowe, Laurent Charlin, and Joelle Pineau.
\newblock A survey of available corpora for building data-driven dialogue
  systems.
\newblock {\em arXiv preprint arXiv:1512.05742}, 2015.

\bibitem{serban2017hierarchical}
Iulian~Vlad Serban, Alessandro Sordoni, Ryan Lowe, Laurent Charlin, Joelle
  Pineau, Aaron~C Courville, and Yoshua Bengio.
\newblock A hierarchical latent variable encoder-decoder model for generating
  dialogues.
\newblock In {\em AAAI}, pages 3295--3301, 2017.

\bibitem{forbes}
Ashley Stahl.
\newblock {10 Ways Enterprise Chatbot Solutions And AI Are Changing The
  Workplace}.
\newblock
  \url{https://www.forbes.com/sites/ashleystahl/2017/07/20/10-ways-enterprise-chatbot-solutions-and-ai-are-changing-the-workplace},
  July 2017.

\bibitem{strzalkowski2010modeling}
Tomek Strzalkowski, George~Aaron Broadwell, Jennifer Stromer-Galley, Samira
  Shaikh, Sarah Taylor, and Nick Webb.
\newblock Modeling socio-cultural phenomena in discourse.
\newblock In {\em Proceedings of the 23rd International Conference on
  Computational Linguistics}, pages 1038--1046. Association for Computational
  Linguistics, 2010.

\bibitem{su2015learning}
Pei-Hao Su, David Vandyke, Milica Gasic, Dongho Kim, Nikola Mrksic, Tsung-Hsien
  Wen, and Steve Young.
\newblock Learning from real users: Rating dialogue success with neural
  networks for reinforcement learning in spoken dialogue systems.
\newblock {\em arXiv preprint arXiv:1508.03386}, 2015.

\bibitem{typke2005survey}
Rainer Typke, Frans Wiering, and Remco~C Veltkamp.
\newblock A survey of music information retrieval systems.
\newblock In {\em Proc. 6th International Conference on Music Information
  Retrieval}, pages 153--160. Queen Mary, University of London, 2005.

\bibitem{walker2012annotated}
Marilyn~A Walker, Grace~I Lin, and Jennifer Sawyer.
\newblock An annotated corpus of film dialogue for learning and characterizing
  character style.
\newblock In {\em LREC}, pages 1373--1378, 2012.

\bibitem{Weiss2012}
Sholom~M. Weiss, Nitin Indurkhya, and Tong Zhang.
\newblock {\em Fundamentals of Predictive Text Mining}.
\newblock Springer Publishing Company, Incorporated, 2012.

\bibitem{wen2016network}
Tsung-Hsien Wen, David Vandyke, Nikola Mrksic, Milica Gasic, Lina~M
  Rojas-Barahona, Pei-Hao Su, Stefan Ultes, and Steve Young.
\newblock A network-based end-to-end trainable task-oriented dialogue system.
\newblock {\em arXiv preprint arXiv:1604.04562}, 2016.

\bibitem{williams2017hybrid}
Jason~D Williams, Kavosh Asadi, and Geoffrey Zweig.
\newblock Hybrid code networks: practical and efficient end-to-end dialog
  control with supervised and reinforcement learning.
\newblock {\em arXiv preprint arXiv:1702.03274}, 2017.

\bibitem{williams2016end}
Jason~D Williams and Geoffrey Zweig.
\newblock End-to-end lstm-based dialog control optimized with supervised and
  reinforcement learning.
\newblock {\em arXiv preprint arXiv:1606.01269}, 2016.

\end{thebibliography}

\end{document}